%% file: acl.tex
\documentclass[11pt]{article}

\usepackage{acl}

\usepackage{times}
\usepackage{latexsym}
\usepackage{graphicx}
\usepackage{tabularx}
\usepackage{booktabs}
\usepackage[whole]{bxcjkjatype}
\usepackage{multirow}
\usepackage{color,soul}

\usepackage{CJKutf8}

\usepackage[T1]{fontenc}

\usepackage[utf8]{inputenc}

\usepackage{microtype}

\title{Chat Translation Error Detection for Assisting Cross-lingual Communications}

\author{Yunmeng Li${}^{1}$ Jun Suzuki${}^{1,3}$ Makoto Morishita${}^{2}$ Kaori Abe${}^{1}$ \\
  {\bf Ryoko Tokuhisa}${}^{1}$ {\bf Ana Brassard}${}^{3,1}$ {\bf Kentaro Inui}${}^{1,3}$\\
  ${}^{1}$Tohoku University
　${}^{2}$NTT
　${}^{3}$RIKEN\\
　\texttt{li.yunmeng.r1@dc.tohoku.ac.jp}
 }

\begin{document}
\maketitle
\begin{abstract}
In this paper, we describe the development of a communication support system that detects erroneous translations to facilitate cross-lingual communications due to the limitations of current machine chat translation methods.
We trained an error detector as the baseline of the system and constructed a new Japanese–English bilingual chat corpus, \textbf{BPersona-chat}, which comprises multi-turn colloquial chats augmented with crowdsourced quality ratings.
The error detector can serve as an encouraging foundation for more advanced erroneous translation detection systems.
\end{abstract}

\section{Introduction}
\label{sec:introduction}

With the expansion of internationalization, there is an increasing demand for cross-lingual communication.
However, while machine translation technologies have demonstrated sound performance in translating documents~\cite{barrault-etal-2019-findings, barrault-etal-2020-findings,nakazawa-etal-2019-overview}, current methods are not always suitable for translating chat~\cite{laubli-etal-2018-machine,toral-etal-2018-attaining,farajian-etal-2020-findings,liang-etal-2021-modeling}.
When a translation system generates erroneous translations, the user may be unable to identify such errors, which can lead to confusion or misunderstanding.
Thus, in this study, we developed a cross-lingual chat assistance system that reduces potential miscommunications by detecting translation errors and notifying the users of their occurrences.
As a critical component of such a system, we propose the erroneous chat translation detection task and conduct an empirical study to model error detection.
An illustration of the baseline task is shown in Figure~\ref{fig:chat-example}.
When the translation system generates a translation that is suspected to be incorrect or not well-connected to the context, we prompt users on the source language side that the translation may be incorrect.
The warning message is expected to encourage users to modify their text into a better translatable form.
Simultaneously, users on the target language side receive the same warning message to indicate that unusual words or passages are likely translation errors.

\input{figures/fig_chat_example}

To support this line of research, we created a new parallel chat corpus, \textbf{BPersona-chat}\footnote{\url{https://github.com/cl-tohoku/BPersona-chat}}, which comprises multi-turn colloquial chats augmented with manually produced gold translations and machine-generated translations with crowdsourced quality labels (\emph{correct} or \emph{erroneous}).
In an experiment, we trained an error detection model that classifies a given translation in a bilingual two-utterance chat as either correct or erroneous (Figure~\ref{fig:chat-example}) and evaluated its performance on the BPersona-chat dataset. 
Our primary contributions are summarized as follows.
(1) We propose the erroneous chat translation detection task.
(2) We construct that BPersona-chat parallel chat corpus.
(3) We trained the error detector, thereby providing a foundation to develop more sophisticated communication support systems.

\section{Task Definition}
\label{sec:task-definition}

As the baseline task, we define a \textit{chat} as a two-utterance colloquial dialog between two humans using different languages.
Here, we focus on predicting whether the second utterance, i.e., the response, was translated correctly.
The preceding context, the translation of the context, the response, and the translated response are input to the error detector.
Then, the detector predicts the translated response using the other utterances as reference data.
The detector then outputs whether the translated response is erroneous.

Figure~\ref{fig:chat-example} shows an example target task of evaluating the Japanese translation of an English utterance.
Here, the Japanese speaker's initial utterance $ja_1$ is translated into $en_1$, and the English speaker's response $en_2$ is translated into $ja_2$.
In this example, the detector is assessing the utterance ``ありがとう。(\textit{Thanks.}),'' which is not an accurate translation of the utterance ``I agree.''
The detector is given the preceding context ($ja_1$, $en_1$, and $en_2$) as reference data to predict whether the translation is both accurate and coherent.
If the detector is predicting the translation $en_2$ of response $ja_2$, the reference data include $en_1$, $ja_1$, and $ja_2$ in the opposite.

\section{Related Work}

\paragraph{Translation quality estimation task}
Our target task is a new setting compared to quality estimation tasks~\cite{specia-etal-2020-findings-wmt,fonseca-etal-2019-findings}, which primarily focus on written text, e.g., Wikipedia articles and Amazon reviews.
In contrast, the target task attempts to detect errors in chat translation systems; thus, we must understand the contexts of casual conversational settings.

\paragraph{Parallel dialog corpus}
There are bilingual dialog corpora, e.g., Business Scene Dialog~\cite{rikters-etal-2019-designing}, which includes business negotiation scenes in both Japanese and English.
However, our task requires data that include cross-lingual colloquial chats with both appropriate and erroneous translations.
To the best of our knowledge, no such dataset exists; thus, we must prepare a new evaluation dataset to evaluate the proposed task.

\section{Evaluation Dataset}
\label{sec:bpersonachat}

To mitigate the construction time and cost, we took advantage of existing chat corpora as a starting point.
We first filtered out inappropriate chats, then asked professional translators to perform utterance-by-utterance translations in consideration of the contexts to acquire correct translation candidates.
In addition, we prepared utterance-by-utterance machine translations, without considering chat contexts to acquire incorrect translation candidates.
Finally, we evaluated the translations to see if they were acceptable chat translations.
The details of each process are described in the following.

\subsection{Base Datasets}

We constructed Japanese–English bidirectional chat translation datasets. 
Specifically, we focused on Persona-chat~\cite{zhang-etal-2018-personalizing} and JPersona-chat~\cite{sugiyama-etal-2021-empirical} as our base datasets.
These datasets contain multiturn chat data in English and Japanese, respectively\footnote{Persona-chat and JPersona-chat are not translations of each other.}.
Each chat was performed between two crowd workers assuming artificial personas.
The speakers discuss a given personality trait, including but not limited to self-introduction, hobby, and others.

\subsection{Filtering Incoherent Data}

\input{tables/tab_incoherent_chat}

A preliminary manual review of the Persona-chat dataset revealed occasionally incoherent chats, e.g., unnatural topic changes or misunderstandings (Table~\ref{tab:incoherent-chat}).
We removed such examples from the dataset by asking crowd workers to flag passages they deemed incoherent
.
Here, we defined ``incoherence'' as questions being ignored, the presence of unnatural topic changes, one speaker not addressing what the other speaker said, responses appearing to be out of order or generally difficult to follow.

We scored each chat according to the workers' answers and selected the top $200$ among $1,500$ chats\footnote{See Appendix~\ref{sec:crowdsourcing} for additional details about the crowdsourcing process.}.
The selected $200$ chats were marked as accurate and coherent by at least seven of the 10 workers.

\subsection{Bilingual Chats with Human Translations}

\input{tables/tab_toprated_chat_with_translation}

To construct a parallel Japanese–English chat corpus, we combined the selected top $200$ top chats ($2,940$ utterances in total) from the Persona-chat dataset and $250$ chats ($2,740$ utterances in total) from the JPersona-chat dataset.
We then translated them into their respective target languages\footnote{We sought consent to translate JPersona-chat with the authors.}.
Here, we commissioned professional translators proficient in Japanese and English to ensure high-quality translations.
We asked the translators to consider both the accuracy of the translation and the coherence of the dialog.
The translators were given information about the personas to help adjust the speaking styles.
As a result, we obtained a parallel corpus of $450$ dialogs ($5,680$ utterances) and their translations, which we refer to as the Bilingual Persona-chat (BPersona-chat) corpus.
Table~\ref{tab:toprated-chat-with-translation} shows a sample from the BPersona-chat corpus.

\subsection{Bilingual Chats with Neural Machine Translation Translations}
\label{sec:nmt-translation}

The task of the error detector is to distinguish between accurate and poor (potentially harmful) translations.
The BPersona-chat corpus provides examples of the former.
Given professionally-translated bilingual chats, we also prepared low-quality alternative translations generated using a machine translation model.
Here, we trained a Transformer-based neural machine translation (NMT) model A on OpenSubtitles2018~\cite{lison-etal-2018-opensubtitles2018}, achieving a BLEU score~\cite{papineni-etal-2002-bleu} of $4.9$ on the BPersona-chat corpus\footnote{Refer to Appendix~\ref{sec:appendix_training_nmt} for additional details about training NMT model A.}.
Note that this BLEU score is relatively low because domain mismatch is possible between OpenSubtitles2018 and the BPersona-chat corpus.
However, it was a preferable setting because we required poor translations to construct our dataset.
In addition, we prepared better translations with a translation model B, which achieved a BLEU score of $26.4$.

\subsection{Human Evaluation of Translations}

To confirm that the alternative translations generated by NMT model A were erroneous to the crowds, we asked crowd workers proficient in both English and Japanese to rate each translation in the chat as either good or bad.
We qualified the workers to ensure they could reach the level of native Japanese, and the level of business and academic English.

The workers rated $5,088$ of NMT model A's $5,680$ ($89.58\%$) translations, $1,718$ of NMT model B's $5,680$ ($30.25\%$) translations, and $597$ of the $5,680$ ($10.51\%$) human translations as bad\footnote{Refer to Appendix\ref{sec:crowdsourcing} for additional details about the crowdsourcing process.}.
Then, each utterance-translation pair was marked as erroneous or correct based on human evaluations.

According to our task settings, an utterance cannot be used as the referenced preceding context if none of it is correct.
Thus, we deleted the $159$ utterances whose human translations, model A's translations, and model B's translations were all erroneous.
As a result, we obtained $2,674$ English utterances with $8,022$ corresponding labeled Japanese translations, where $3,406$ of the translations were labeled as erroneous, and the remaining $4,616$ translations were labeled as correct.
In addition, we obtained $2,397$ Japanese utterances with $7,190$ corresponding labeled English translations, where $3,096$ translations were labeled as erroneous, and $4,094$ were labeled as correct.
These labeled data were used to evaluate the error detector in our subsequent experiments.

\section{Baseline Error Detecting Classifier}
\label{sec:classifier}

As a baseline approach, we trained and evaluated a binary BERT-based~\cite{devlin-etal-2018-bert, wolf-etal-2020-transformers} classifier as the error detector\footnote{Refer to Appendix\ref{sec:appendix_clasification_model} for additional details about training this classification model.}.
Here, the input was structured as ``$ja_1$\texttt{[SEP]}$en_1$\texttt{[SEP]}$en_2$\texttt{[SEP]}$ja_2$'' to predict the Japanese translation $ja_2$ of the corresponding source utterance $en_2$.
The input was structured as ``$en_1$\texttt{[SEP]}$ja_1$\texttt{[SEP]}$ja_2$\texttt{[SEP]}$en_2$'' to predict the translation $en_2$ of the corresponding source utterance $ja_2$ in the opposite translating direction\footnote{\texttt{[SEP]} was used to indicate different utterances, \texttt{[CLS]} was used to indicate the beginning of the data and \texttt{[PAD]} was used as the padding token.}.
Similar to the original experimental settings for BERT, we applied the SoftMax function to the classification result to obtain the final prediction.

We used the OpenSubtitles2018 dataset for training with approximately one million utterances.
Here, we generated negative samples with the low-quality translation model A (Section~\ref{sec:nmt-translation}), and we fine-tuned the multilingual BERT model provided by HuggingFace\footnote{\url{https://huggingface.co/}} to construct the error detector for both the English-to-Japanese and Japanese-to-English directions.

\section{Experiments}

In this section, we report on our trial of the chat translation error detection task (Section~\ref{sec:task-definition}) using the model described in Section~\ref{sec:classifier}.
The task was evaluated with the dataset described in Section ~\ref{sec:bpersonachat}.

\subsection{Evaluation Metrics}

\paragraph{Majority class and minority class classifiers}
To confirm that the error detector is not simply making lucky guesses, we calculated the accuracy of the majority class classifier, the minority class classifier, and the error detector.
Note that the majority class of the data is the correct translation, and the minority class is the erroneous translation.

\paragraph{F-score, precision and recall}
We evaluated the performance of the error detector according to the F-score (\textbf{F}).
We also show the precision (\textbf{Pre}) and recall (\textbf{Rec}) values for reference.
The truth (T) is set as the erroneous translation, and the positive case (P) is detecting the erroneous translation.

\paragraph{Confusion matrix}
To evaluated the performance of the error detector on different types of translations, we provide confusion matrices according to whether the translation was translated by the human translator, NMT model A, or NMT model B.


\input{tables/tab_major_minor}
\input{tables/tab_baseline_score}
\input{tables/tab_confusion_matrix}

\subsection{Results}
\label{sec:results}

The results demonstrate that the error detector is capable for classifying erroneous translations in chats.
According to the accuracy values given in Table~\ref{tab:major-minor}, we conclude that the error detector gained higher performance compared to the majority and minority classifiers.
The results suggest that the current method can solve the task without relying on lucky guesses.
According to the F-score, precision, and recall values shown in Table~\ref{tab:baseline-score}, the error detector could identify erroneous translations in the BPersona-chat dataset.

However, although the detector could distinguish translations with terrible translation or coherence issues, it could not successfully identify errors that were not obvious.
The confusion matrix of the results is shown in Table~\ref{tab:confusion-matrix}, where the row headers are the actual annotations, and the column headers are the labels predicted by the detector.
As can be seen, the error detector did not perform well when attempting to predict the translations generated by the high-quality NMT model B.
Here, the detector labeled more than half of the erroneous translations generated by NMT model B as correct.
One possible reason for this is that the detector was trained on a dataset whose erroneous examples were generated by model A, which generated low-quality translations.

\input{tables/tab_chat_high_sent_label_zero}

To compare the error detector with the traditional BLEU calculation, we calculated the sentence-BLEU score of each utterance in the BPersona-chat dataset using the method provided by NLTK~\cite{bird-etal-2009-nltk}.
The results demonstrate that the detector can help distinguish an erroneous translation even when the translation has a high BLEU score. Table~\ref{tab:chat-high-sent-label-zero} shows an example of a translation $en_2$ with a high sentence-BLEU score but incorrectly translated the Japanese word ``米'' into ``America'' rather than ``rice''.
We found that the detector helped distinguish this case as erroneous, as was expected.

\subsection{Quality of the Evaluation Dataset}

The reason a considerably high score was obtained on the NMT model A's translations is not entirely straightforward.
Note that we trained the classification model on OpenSubtitles2018, which has a different distribution from BPersona-chat.
This means that the training was performed using out-of-domain data.
One potential reason for the high performance may be attributed to the nature of the automatically generated translations.
As with the experimental results described in Section~\ref{sec:results}, it was difficult for the detector to distinguish the good translations generated using the high-quality NMT model B.
To improve performance, it is important to clarify the exact issue with the erroneous translation.

\section{Discussions and Future Work}

In this paper, we have proposed the chat translation error detection task to assist cross-lingual communication.
For this purpose, we constructed a parallel Japanese–English chat corpus as the backbone for evaluation, including high-quality and low-quality translations augmented with crowdsourced quality ratings.
We trained the error detector to identify erroneous translations, and the detector could help detect the erroneous translations in chat.

While this is the first trial to realize a cross-lingual chat assistance system, we hope to promote research to complete the chat translation assistance system in the future, and we aim to advance the detector's ability to indicate the translation's critical error possibility.
This will allow speakers to focus on translations with high error rates.
In addition, we hope to identify specific errors in the translations for users.
To achieve this goal, we would like to refine the BPersona-chat dataset with multiple labels corresponding to different translation errors.
The binary classification model would also be improved into multi-label, which would enable the error detector to analyze concrete problems.
Thus, we would be able to identify the exact error in the current speech for revisions.
We will also consider providing translation suggestions as reference information to help users modify.

When both parties cannot understand each other's language, the advanced error detecting system is expected to alert them of possible errors and guide them to modify their texts, thereby reducing translation problems in multilingual chats.
Finding a balance between coherence and accuracy is always difficult in chat translation.
However, we believe that advancing and refining the error detector and the corresponding dataset will help us identify and solve specific problems in chat translation systems.

\section*{Acknowledgements}
This work was supported by JST (the establishment of university fellowships towards the creation of science technology innovation) Grant Number JPMJFS2102, JST CREST Grant Number JPMJCR20D2 and JST Moonshot R\&D Grant Number JPMJMS2011 (fundamental research).
The crowdsourcing was supported by Amazon Mechanical Turk (\url{https://www.mturk.com/}) and Crowdworks (\url{https://crowdworks.jp/}).

\bibliography{anthology}
\bibliographystyle{acl_natbib}

\clearpage
\appendix

\section{Settings of Machine Translation Model}
\label{sec:appendix_training_nmt}

\begin{table}[tb]
  \centering
  \small
  \tabcolsep=1pt
  \begin{tabular}{lp{40mm}}
  \toprule
    Architecture            &  2-to-2 Transformer~\cite{vaswani:2017:NIPS,tiedemann-scherrer-2017-neural} \\
    Enc-Dec layers & 6 \\
    Attention heads & 8 \\
    Word-embedding dimension & 512 \\
    Feed-forward dimension & 2,048 \\
    Share all embeddings & True \\
    Optimizer              &   Adam ($\beta_{1}=0.9, \beta_{2}=0.98, \epsilon=1\times10^{-8}$)~\cite{kingma:2015:ICLR}  \\
    Learning rate schedule &   Inverse square root decay     \\
    Warmup steps           &   4,000  \\
    Max learning rate      &   0.001  \\
    Initial Learning Rate  &   1e-07  \\
    Dropout                &   0.3~\cite{JMLR:v15:srivastava14a}  \\
    Label smoothing        &   $\epsilon_{ls}=0.1$~\cite{szegedy:2016:rethinking}     \\
    Mini-batch size        &   8,000 tokens~\cite{ott:2018:scaling}\\
    Number of epochs      &   20 \\
    Averaging              &   Save checkpoint for every 5000 iterations and take an average of last five checkpoints \\
    Beam size              &   6 with length normalization~\cite{wu16google}\\
    Implementation         &   \texttt{fairseq}~\cite{ott:2019:fairseq} \\
  \bottomrule
  \end{tabular}
  \caption{List of hyper-parameters for training the NMT model A}
  \label{tab:hyper-parameter-nmt}
\end{table}

This section describes the details of the training neural machine translation model.
Firstly, we tokenized the corpus into subwords with BPE~\cite{sennrich-etal-2016-neural}.
We set the vocabulary size to 32,000.
Then we trained the 2-to-2 Transformer-based NMT model A~\cite{tiedemann-scherrer-2017-neural}, which outputs two consecutive given two input sentences to consider larger contexts.
Table~\ref{tab:hyper-parameter-nmt} shows the list of hyper-parameters.

\section{Settings of Classification Model}
\label{sec:appendix_clasification_model}

\begin{table}[tb]
  \centering
  \small
  \tabcolsep=1pt
  \begin{tabular}{lp{40mm}}
  \toprule
    Architecture            &  BERT (base)~\cite{devlin-etal-2018-bert} \\
    Optimizer              &   Adam ($\beta_{1}=0.9, \beta_{2}=0.98, \epsilon=1\times10^{-8}$, weight decay=0.01)~\cite{kingma:2015:ICLR}  \\
    Learning rate schedule &   Inverse square root decay     \\
    Max learning rate      &   0.001  \\
    Mini-batch size        &   16 samples\\
    Number of epochs      &   1 \\
    Implementation         &   \texttt{transformers}~\cite{wolf-etal-2020-transformers} \\
  \bottomrule
  \end{tabular}
  \caption{List of hyper-parameters for training the classification model}
  \label{tab:hyper-parameter-classifier}
\end{table}

This section describes the details of the training classification model.
Table~\ref{tab:hyper-parameter-classifier} shows the list of hyper-parameters.

\section{Details of Crowd-sourcing Tasks}
\label{sec:crowdsourcing}

\subsection{Filtering Persona-chat}

We asked crowd workers on Amazon Mechanical Turk (\url{https://requester.mturk.com/}) to filter out incoherent data in Persona-chat.
Here, we defined a chat as ``incoherent'' if:
\begin{itemize}
    \item questions being ignored;
    \item the presence of unnatural topic changes;
    \item one is not addressing what the other said;
    \item responses seeming out of order;
    \item or being hard to follow in general.
\end{itemize}
Workers were instructed to disregard minor issues such as typos and focus on the general flow.

In the full round, we selected $1,500$ chats from Persona-chat.
Each crowd worker was tasked to rate $5$ chats at a time, and each chat was rated by $10$ different workers. 
Eligible workers were selected with a preliminary qualification round.

\subsection{Rating Translations}

We asked crowd workers on Crowdworks (\url{https://crowdworks.jp/}) to label the human translation and the NMT translation in BPersona-chat as low-quality or high-quality.
In the task, we defined a translation as bad if:
\begin{itemize}
    \item the translation is incorrect; 
    \item parts of the source chat are lost; 
    \item there are serious grammatical or spelling errors that interfere with understanding; 
    \item the person's speaking style changes from the past utterance; 
    \item the translation is meaningless or incomprehensible;
    \item or the translation is terrible in general.
\end{itemize}
Workers worked on files in which one file included one complete chat; therefore, they could check the context and rate each utterance of the conversation.

To the limited number of workers, in the full round, crowd workers were tasked to rate around $50$ to $300$ chats in two weeks. 
Eligible workers were selected with a preliminary qualification round.

\end{document}

%% file: figures/fig_chat_example.tex
\begin{figure}[t]
\centering
\includegraphics[width=\columnwidth]{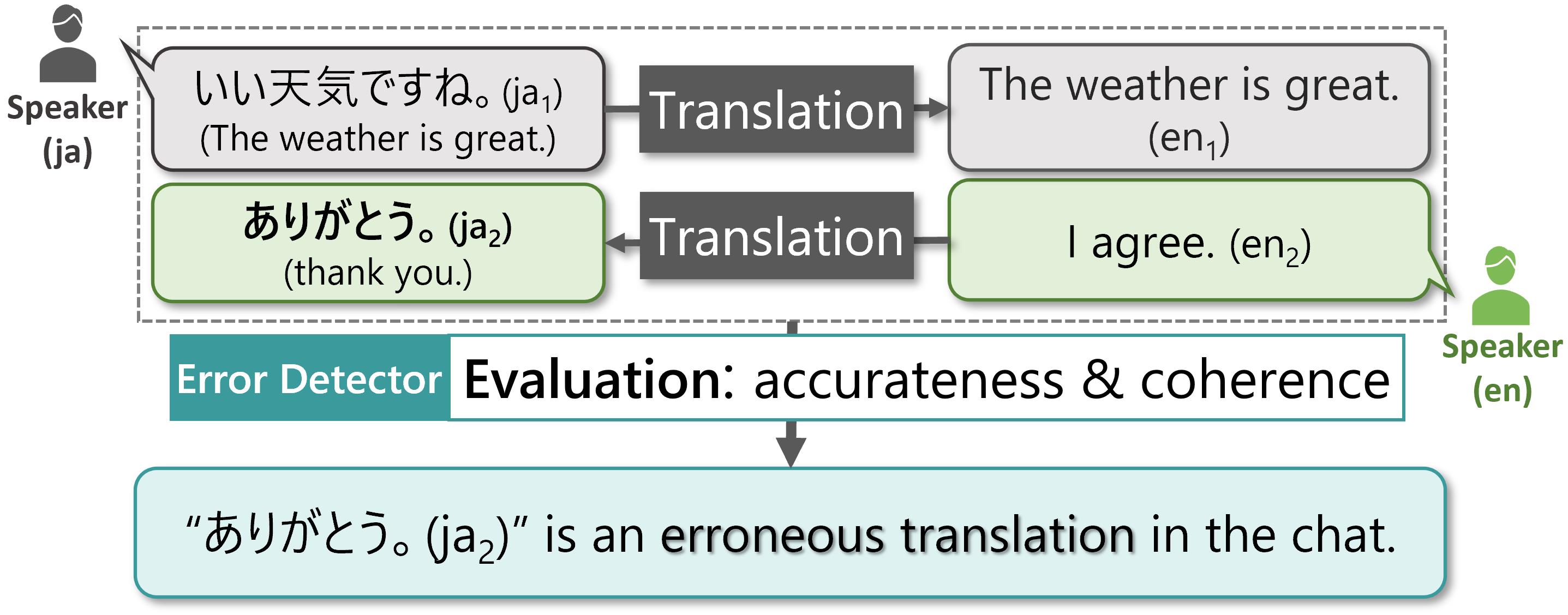}
\caption{Illustration of the error detector predicting erroneous translations. The detector evaluates whether translation $ja_2$ is accurate and coherent in the chat.}
\label{fig:chat-example}
\vspace{-1mm}
\end{figure}

%% file: tables/tab_incoherent_chat.tex
\begin{table}[t]
\centering
\small
\begin{tabular}{lp{5.5cm}}
\toprule
\textbf{Speaker}  & \textbf{Utterance} \\ \midrule
person 1    & I do not like carrots. I throw them away. \\ 
person 2    & really. I can sing pitch perfect. \textit{(incoherent: carrots → sing)} \\ 
person 1    & I also cook, and I ride my bike to work. \textit{(incoherent: sing → ride)} \\ 
person 2    & great! I had won an award for spelling bee. \textit{(incoherent: ride → spelling)}  \\ \bottomrule
\end{tabular}
\caption{Example of incoherent chat from Persona-chat.}
\label{tab:incoherent-chat}
\end{table}

%% file: tables/tab_toprated_chat_with_translation.tex
\begin{table*}[t] 
\centering 
\small 
\begin{tabular}{lp{6.5cm}p{6.5cm}} 
\toprule \textbf{Speaker} & \textbf{Original utterance in Perosona-chat (en)} & \textbf{Translation by professional translators (ja)} \\ \midrule 
person 1 &Good evening, how has your day been? &こんばんは、今日はどうだった？\\ 
person 2 &It was good I met up with some friends to larp &よかったよ、ライブRPGで友達と集まった。\\ 
person 1 &I wish I had time for that, working 40 hours in a bank is killing me. &そんな時間があればなあ、銀行で４０時間勤務は死にそうだよ。\\ 
person 2 & ... ... & ... ... \\ \bottomrule 
\end{tabular} 
\caption{Example of the top 200 coherent chats from the Persona-chat dataset as rated by crowdsourcing workers and translated to Japanese by professional translators.} 
\label{tab:toprated-chat-with-translation} 
\end{table*}

%% file: tables/tab_major_minor.tex
\begin{table}[]
\centering
\small
\begin{tabular}{@{}ccllcll@{}}
\toprule
\textbf{}               & \multicolumn{3}{c}{\textbf{ja→en}} & \multicolumn{3}{c}{\textbf{en→ja}} \\ \midrule
\textbf{Majority class} & \multicolumn{3}{c}{56.94}          & \multicolumn{3}{c}{57.54}          \\
\textbf{Minority class} & \multicolumn{3}{c}{43.06}          & \multicolumn{3}{c}{42.46}          \\
\textbf{Error detector} & \multicolumn{3}{c}{76.27}          & \multicolumn{3}{c}{77.06}          \\ \bottomrule
\end{tabular}
\caption{Accuracy of the majority class classifier, minority class classifier, and error detector.}
\label{tab:major-minor}
\end{table}

%% file: tables/tab_baseline_score.tex
\begin{table}[t]
\centering
\small
\tabcolsep 3pt
\begin{tabular}{@{}lllllll@{}}
\toprule
& \multicolumn{3}{c}{\textbf{ja $\rightarrow$ en}} & \multicolumn{3}{c}{\textbf{en $\rightarrow$ ja}} \\
&
  \multicolumn{1}{c}{\textbf{F}} &
  \multicolumn{1}{c}{(\textbf{Pre}} &
  \multicolumn{1}{c}{\textbf{Rec})} &
  \multicolumn{1}{c}{\textbf{F}} &
  \multicolumn{1}{c}{(\textbf{Pre}} &
  \multicolumn{1}{c}{\textbf{Rec})} \\ \midrule
\textbf{Error detector} & 73.30 & (71.10 & 75.65) & 75.03 & (69.75 & 81.18) \\
\bottomrule
\end{tabular}
\caption{F-score, precision, and recall of the error detector on BPersona-chat dataset.}
\label{tab:baseline-score}
\end{table}

%% file: tables/tab_confusion_matrix.tex
\begin{table*}[t]
\centering
\small
\begin{tabular}{@{}ccccccccc@{}}
\toprule
\multicolumn{9}{c}{\textbf{ja→en}} \\ \midrule
\multicolumn{3}{c}{\textbf{Human}} & \multicolumn{3}{c}{\textbf{NMT model A (low-quality)}} & \multicolumn{3}{c}{\textbf{NMT model B (high-quality)}} \\ \midrule
\multicolumn{1}{l}{} & \textbf{Correct} & \textbf{Erroneous} & \multicolumn{1}{l}{\textbf{}} & \textbf{Correct} & \textbf{Erroneous} & \multicolumn{1}{l}{\textbf{}} & \textbf{Correct} & \textbf{Erroneous} \\
\textbf{Correct} & 1879 & 207 & \textbf{Correct} & 11 & 155 & \textbf{Correct} & 1252 & 590 \\
\textbf{Erroneous} & 290 & 21 & \textbf{Erroneous} & 90 & 2140 & \textbf{Erroneous} & 374 & 181 \\ \midrule
\multicolumn{9}{c}{\textbf{en→ja}} \\ \midrule
\multicolumn{3}{c}{\textbf{Human}} & \multicolumn{3}{c}{\textbf{NMT model A (low-quality)}} & \multicolumn{3}{c}{\textbf{NMT model B (high-quality)}} \\ \midrule
\multicolumn{1}{l}{} & \textbf{Correct} & \textbf{Erroneous} & \multicolumn{1}{l}{} & \textbf{Correct} & \textbf{Erroneous} & \multicolumn{1}{l}{} & \textbf{Correct} & \textbf{Erroneous} \\
\textbf{Correct} & 2406 & 176 & \textbf{Correct} & 6 & 265 & \textbf{Correct} & 1005 & 758 \\
\textbf{Erroneous} & 83 & 9 & \textbf{Erroneous} & 53 & 2350 & \textbf{Erroneous} & 505 & 406 \\ \bottomrule
\end{tabular}
\caption{Confusion matrix of the error detector on BPersona-chat data (row headers are the actual annotations, and column headers are the prediction made by the detector).}
\label{tab:confusion-matrix}
\end{table*}

%% file: tables/tab_chat_high_sent_label_zero.tex
\begin{table}[t]
\centering
\small
\begin{tabular}{@{}ll@{}}
\toprule
$en_1$ (context)            & What did you have for dinner? \\ 
$ja_1$                      & 晩ご飯に何を食べましたか？ \\ 
$ja_2$ (source)             & 晩ご飯に米を食べました。 \\ 
\textbf{$en_2$ (translation)} & \textbf{I had America as my dinner.} \\ 
\textit{(reference)}        & \textit{(I had rice as my dinner.)} \\ \midrule
sentence-BLEU               & 72.7 (compared to the reference) \\ 
classifier's prediction     & erroneous \\ \bottomrule
\end{tabular}
\caption{Example where the error detector successfully predicted the erroneous translation $en_2$ even though it had a high sentence-BLEU score.}
\label{tab:chat-high-sent-label-zero}
\end{table}

%% file: acl.bbl
\begin{thebibliography}{26}
\expandafter\ifx\csname natexlab\endcsname\relax\def\natexlab#1{#1}\fi

\bibitem[{Barrault et~al.(2020)Barrault, Biesialska, Bojar, Costa-juss{\`a},
  Federmann, Graham, Grundkiewicz, Haddow, Huck, Joanis, Kocmi, Koehn, Lo,
  Ljube{\v{s}}i{\'c}, Monz, Morishita, Nagata, Nakazawa, Pal, Post, and
  Zampieri}]{barrault-etal-2020-findings}
Lo{\"\i}c Barrault, Magdalena Biesialska, Ond{\v{r}}ej Bojar, Marta~R.
  Costa-juss{\`a}, Christian Federmann, Yvette Graham, Roman Grundkiewicz,
  Barry Haddow, Matthias Huck, Eric Joanis, Tom Kocmi, Philipp Koehn, Chi-kiu
  Lo, Nikola Ljube{\v{s}}i{\'c}, Christof Monz, Makoto Morishita, Masaaki
  Nagata, Toshiaki Nakazawa, Santanu Pal, Matt Post, and Marcos Zampieri. 2020.
\newblock Findings of the 2020 conference on machine translation ({WMT}20).
\newblock In \emph{Proceedings of the Fifth Conference on Machine Translation},
  pages 1--55.

\bibitem[{Barrault et~al.(2019)Barrault, Bojar, Costa-juss{\`a}, Federmann,
  Fishel, Graham, Haddow, Huck, Koehn, Malmasi, Monz, M{\"u}ller, Pal, Post,
  and Zampieri}]{barrault-etal-2019-findings}
Lo{\"\i}c Barrault, Ond{\v{r}}ej Bojar, Marta~R. Costa-juss{\`a}, Christian
  Federmann, Mark Fishel, Yvette Graham, Barry Haddow, Matthias Huck, Philipp
  Koehn, Shervin Malmasi, Christof Monz, Mathias M{\"u}ller, Santanu Pal, Matt
  Post, and Marcos Zampieri. 2019.
\newblock Findings of the 2019 conference on machine translation ({WMT}19).
\newblock In \emph{Proceedings of the Fourth Conference on Machine Translation
  (Volume 2: Shared Task Papers, Day 1)}, pages 1--61.

\bibitem[{Bird et~al.(2009)Bird, Klein, and Loper}]{bird-etal-2009-nltk}
Steven Bird, Ewan Klein, and Edward Loper. 2009.
\newblock \emph{Natural language processing with Python: analyzing text with
  the natural language toolkit}.
\newblock " O'Reilly Media, Inc.".

\bibitem[{Devlin et~al.(2019)Devlin, Chang, Lee, and
  Toutanova}]{devlin-etal-2018-bert}
Jacob Devlin, Ming-Wei Chang, Kenton Lee, and Kristina Toutanova. 2019.
\newblock {BERT}: Pre-training of deep bidirectional transformers for language
  understanding.
\newblock In \emph{Proceedings of the 2019 Conference of the North {A}merican
  Chapter of the Association for Computational Linguistics: Human Language
  Technologies, Volume 1 (Long and Short Papers)}, pages 4171--4186.

\bibitem[{Farajian et~al.(2020)Farajian, Lopes, Martins, Maruf, and
  Haffari}]{farajian-etal-2020-findings}
M.~Amin Farajian, Ant{\'o}nio~V. Lopes, Andr{\'e} F.~T. Martins, Sameen Maruf,
  and Gholamreza Haffari. 2020.
\newblock Findings of the {WMT} 2020 shared task on chat translation.
\newblock In \emph{Proceedings of the Fifth Conference on Machine Translation},
  pages 65--75.

\bibitem[{Fonseca et~al.(2019)Fonseca, Yankovskaya, Martins, Fishel, and
  Federmann}]{fonseca-etal-2019-findings}
Erick Fonseca, Lisa Yankovskaya, Andr{\'e} F.~T. Martins, Mark Fishel, and
  Christian Federmann. 2019.
\newblock \href {https://aclanthology.org/W19-5401} {Findings of the {WMT} 2019
  shared tasks on quality estimation}.
\newblock In \emph{Proceedings of the Fourth Conference on Machine Translation
  (Volume 3: Shared Task Papers, Day 2)}, pages 1--10.

\bibitem[{Kingma and Ba(2015)}]{kingma:2015:ICLR}
Diederik Kingma and Jimmy Ba. 2015.
\newblock \href {https://arxiv.org/abs/1412.6980} {{Adam}: A method for
  stochastic optimization}.
\newblock In \emph{Proceedings of the 3rd International Conference on Learning
  Representations (ICLR 2015)}.

\bibitem[{L{\"a}ubli et~al.(2018)L{\"a}ubli, Sennrich, and
  Volk}]{laubli-etal-2018-machine}
Samuel L{\"a}ubli, Rico Sennrich, and Martin Volk. 2018.
\newblock Has machine translation achieved human parity? a case for
  document-level evaluation.
\newblock In \emph{Proceedings of the 2018 Conference on Empirical Methods in
  Natural Language Processing}, pages 4791--4796.

\bibitem[{Liang et~al.(2021)Liang, Meng, Chen, Xu, and
  Zhou}]{liang-etal-2021-modeling}
Yunlong Liang, Fandong Meng, Yufeng Chen, Jinan Xu, and Jie Zhou. 2021.
\newblock \href {http://arxiv.org/abs/2107.11164} {Modeling bilingual
  conversational characteristics for neural chat translation}.

\bibitem[{Lison et~al.(2018)Lison, Tiedemann, and
  Kouylekov}]{lison-etal-2018-opensubtitles2018}
Pierre Lison, J{\"o}rg Tiedemann, and Milen Kouylekov. 2018.
\newblock {O}pen{S}ubtitles2018: Statistical rescoring of sentence alignments
  in large, noisy parallel corpora.
\newblock In \emph{Proceedings of the Eleventh International Conference on
  Language Resources and Evaluation ({LREC} 2018)}, pages 1742--1748.

\bibitem[{Nakazawa et~al.(2019)Nakazawa, Doi, Higashiyama, Ding, Dabre, Mino,
  Goto, Pa, Kunchukuttan, Oda, Parida, Bojar, and
  Kurohashi}]{nakazawa-etal-2019-overview}
Toshiaki Nakazawa, Nobushige Doi, Shohei Higashiyama, Chenchen Ding, Raj Dabre,
  Hideya Mino, Isao Goto, Win~Pa Pa, Anoop Kunchukuttan, Yusuke Oda,
  Shantipriya Parida, Ond{\v{r}}ej Bojar, and Sadao Kurohashi. 2019.
\newblock \href {https://aclanthology.org/D19-5201} {Overview of the 6th
  workshop on {A}sian translation}.
\newblock In \emph{Proceedings of the 6th Workshop on Asian Translation}, pages
  1--35.

\bibitem[{Ott et~al.(2019)Ott, Edunov, Baevski, Fan, Gross, Ng, Grangier, and
  Auli}]{ott:2019:fairseq}
Myle Ott, Sergey Edunov, Alexei Baevski, Angela Fan, Sam Gross, Nathan Ng,
  David Grangier, and Michael Auli. 2019.
\newblock \href {https://doi.org/10.18653/v1/N19-4009} {{fairseq: A Fast,
  Extensible Toolkit for Sequence Modeling}}.
\newblock In \emph{Proceedings of the 2019 Conference of the North {A}merican
  Chapter of the Association for Computational Linguistics (Demonstrations)},
  pages 48--53.

\bibitem[{Ott et~al.(2018)Ott, Edunov, Grangier, and Auli}]{ott:2018:scaling}
Myle Ott, Sergey Edunov, David Grangier, and Michael Auli. 2018.
\newblock \href {https://doi.org/10.18653/v1/W18-6301} {{Scaling Neural Machine
  Translation}}.
\newblock In \emph{Proceedings of the Third Conference on Machine Translation
  (WMT 2018)}, pages 1--9.

\bibitem[{Papineni et~al.(2002)Papineni, Roukos, Ward, and
  Zhu}]{papineni-etal-2002-bleu}
Kishore Papineni, Salim Roukos, Todd Ward, and Wei-Jing Zhu. 2002.
\newblock \href {https://doi.org/10.3115/1073083.1073135} {{B}leu: a method for
  automatic evaluation of machine translation}.
\newblock In \emph{Proceedings of the 40th Annual Meeting of the Association
  for Computational Linguistics}, pages 311--318, Philadelphia, Pennsylvania,
  USA. Association for Computational Linguistics.

\bibitem[{Rikters et~al.(2019)Rikters, Ri, Li, and
  Nakazawa}]{rikters-etal-2019-designing}
Mat{\=\i}ss Rikters, Ryokan Ri, Tong Li, and Toshiaki Nakazawa. 2019.
\newblock \href {https://doi.org/10.18653/v1/D19-5204} {Designing the business
  conversation corpus}.
\newblock In \emph{Proceedings of the 6th Workshop on Asian Translation}, pages
  54--61, Hong Kong, China. Association for Computational Linguistics.

\bibitem[{Sennrich et~al.(2016)Sennrich, Haddow, and
  Birch}]{sennrich-etal-2016-neural}
Rico Sennrich, Barry Haddow, and Alexandra Birch. 2016.
\newblock \href {https://doi.org/10.18653/v1/P16-1162} {Neural machine
  translation of rare words with subword units}.
\newblock In \emph{Proceedings of the 54th Annual Meeting of the Association
  for Computational Linguistics (Volume 1: Long Papers)}, pages 1715--1725,
  Berlin, Germany. Association for Computational Linguistics.

\bibitem[{Specia et~al.(2020)Specia, Blain, Fomicheva, Fonseca, Chaudhary,
  Guzm{\'a}n, and Martins}]{specia-etal-2020-findings-wmt}
Lucia Specia, Fr{\'e}d{\'e}ric Blain, Marina Fomicheva, Erick Fonseca, Vishrav
  Chaudhary, Francisco Guzm{\'a}n, and Andr{\'e} F.~T. Martins. 2020.
\newblock \href {https://aclanthology.org/2020.wmt-1.79} {Findings of the {WMT}
  2020 shared task on quality estimation}.
\newblock In \emph{Proceedings of the Fifth Conference on Machine Translation},
  pages 743--764.

\bibitem[{Srivastava et~al.(2014)Srivastava, Hinton, Krizhevsky, Sutskever, and
  Salakhutdinov}]{JMLR:v15:srivastava14a}
Nitish Srivastava, Geoffrey Hinton, Alex Krizhevsky, Ilya Sutskever, and Ruslan
  Salakhutdinov. 2014.
\newblock \href {http://jmlr.org/papers/v15/srivastava14a.html} {Dropout: A
  simple way to prevent neural networks from overfitting}.
\newblock \emph{Journal of Machine Learning Research}, 15(56):1929--1958.

\bibitem[{Sugiyama et~al.(2021)Sugiyama, Mizukami, Arimoto, Narimatsu, Chiba,
  Nakajima, and Meguro}]{sugiyama-etal-2021-empirical}
Hiroaki Sugiyama, Masahiro Mizukami, Tsunehiro Arimoto, Hiromi Narimatsu, Yuya
  Chiba, Hideharu Nakajima, and Toyomi Meguro. 2021.
\newblock \href {http://arxiv.org/abs/2109.05217} {Empirical analysis of
  training strategies of transformer-based japanese chit-chat systems}.

\bibitem[{Szegedy et~al.(2016)Szegedy, Vanhoucke, Ioffe, Shlens, and
  Wojna}]{szegedy:2016:rethinking}
Christian Szegedy, Vincent Vanhoucke, Sergey Ioffe, Jon Shlens, and Zbigniew
  Wojna. 2016.
\newblock \href
  {https://www.cv-foundation.org/openaccess/content_cvpr_2016/papers/Szegedy_Rethinking_the_Inception_CVPR_2016_paper.pdf}
  {{Rethinking the Inception Architecture for Computer Vision}}.
\newblock In \emph{2016 IEEE Conference on Computer Vision and Pattern
  Recognition (CVPR 2016)}, pages 2818--2826.

\bibitem[{Tiedemann and Scherrer(2017)}]{tiedemann-scherrer-2017-neural}
J{\"o}rg Tiedemann and Yves Scherrer. 2017.
\newblock Neural machine translation with extended context.
\newblock In \emph{Proceedings of the Third Workshop on Discourse in Machine
  Translation}, pages 82--92.

\bibitem[{Toral et~al.(2018)Toral, Castilho, Hu, and
  Way}]{toral-etal-2018-attaining}
Antonio Toral, Sheila Castilho, Ke~Hu, and Andy Way. 2018.
\newblock Attaining the unattainable? reassessing claims of human parity in
  neural machine translation.
\newblock In \emph{Proceedings of the Third Conference on Machine Translation:
  Research Papers}, pages 113--123.

\bibitem[{Vaswani et~al.(2017)Vaswani, Shazeer, Parmar, Uszkoreit, Jones,
  Gomez, Kaiser, and Polosukhin}]{vaswani:2017:NIPS}
Ashish Vaswani, Noam Shazeer, Niki Parmar, Jakob Uszkoreit, Llion Jones,
  Aidan~N Gomez, {\L}ukasz Kaiser, and Illia Polosukhin. 2017.
\newblock \href {https://papers.nips.cc/paper/7181-attention-is-all-you-need}
  {{Attention Is All You Need}}.
\newblock In \emph{Advances in Neural Information Processing Systems 31 (NIPS
  2017)}, pages 5998--6008.

\bibitem[{Wolf et~al.(2020)Wolf, Debut, Sanh, Chaumond, Delangue, Moi, Cistac,
  Rault, Louf, Funtowicz, Davison, Shleifer, von Platen, Ma, Jernite, Plu, Xu,
  Scao, Gugger, Drame, Lhoest, and Rush}]{wolf-etal-2020-transformers}
Thomas Wolf, Lysandre Debut, Victor Sanh, Julien Chaumond, Clement Delangue,
  Anthony Moi, Pierric Cistac, Tim Rault, Rémi Louf, Morgan Funtowicz, Joe
  Davison, Sam Shleifer, Patrick von Platen, Clara Ma, Yacine Jernite, Julien
  Plu, Canwen Xu, Teven~Le Scao, Sylvain Gugger, Mariama Drame, Quentin Lhoest,
  and Alexander~M. Rush. 2020.
\newblock Transformers: State-of-the-art natural language processing.
\newblock In \emph{Proceedings of the 2020 Conference on Empirical Methods in
  Natural Language Processing: System Demonstrations}, pages 38--45.

\bibitem[{Wu et~al.(2016)Wu, Schuster, Chen, Le, Norouzi, Macherey, Krikun,
  Cao, Gao, Macherey, Klingner, Shah, Johnson, Liu, Łukasz Kaiser, Gouws,
  Kato, Kudo, Kazawa, Stevens, Kurian, Patil, Wang, Young, Smith, Riesa,
  Rudnick, Vinyals, Corrado, Hughes, and Dean}]{wu16google}
Yonghui Wu, Mike Schuster, Zhifeng Chen, Quoc~V. Le, Mohammad Norouzi, Wolfgang
  Macherey, Maxim Krikun, Yuan Cao, Qin Gao, Klaus Macherey, Jeff Klingner,
  Apurva Shah, Melvin Johnson, Xiaobing Liu, Łukasz Kaiser, Stephan Gouws,
  Yoshikiyo Kato, Taku Kudo, Hideto Kazawa, Keith Stevens, George Kurian,
  Nishant Patil, Wei Wang, Cliff Young, Jason Smith, Jason Riesa, Alex Rudnick,
  Oriol Vinyals, Greg Corrado, Macduff Hughes, and Jeffrey Dean. 2016.
\newblock \href {http://arxiv.org/abs/1609.08144} {Google's neural machine
  translation system: Bridging the gap between human and machine translation}.
\newblock \emph{CoRR}, abs/1609.08144.

\bibitem[{Zhang et~al.(2018)Zhang, Dinan, Urbanek, Szlam, Kiela, and
  Weston}]{zhang-etal-2018-personalizing}
Saizheng Zhang, Emily Dinan, Jack Urbanek, Arthur Szlam, Douwe Kiela, and Jason
  Weston. 2018.
\newblock \href {http://arxiv.org/abs/1801.07243} {Personalizing dialogue
  agents: I have a dog, do you have pets too?}

\end{thebibliography}
